\documentclass{article}
\usepackage{arxiv}

\usepackage{graphicx}
\usepackage[misc]{ifsym}
\usepackage{multicol}
\usepackage{multirow}
\usepackage{subfigure}

\usepackage[utf8]{inputenc} 
\usepackage[T1]{fontenc}    
\usepackage{hyperref}       
\usepackage{url}            
\usepackage{booktabs}       
\usepackage{amsfonts}       
\usepackage{nicefrac}       
\usepackage{microtype}      
\usepackage[square,numbers]{natbib}
\usepackage{doi}

\begin{document}



\title{A-star path planning simulation for UAS Traffic Management (UTM) application}




\author{ 
{Carlos Augusto P\"otter Neto}\\
	Master of Engineering\\
	Aeronautics Institute of Technology (ITA)\\
	São José dos Campos, SP, 12228-900, Brazil \\
	\texttt{carlospottern@gmail.com} \\
	\And

\href{https://orcid.org/0000-0003-1940-8295}{\includegraphics[scale=0.06]{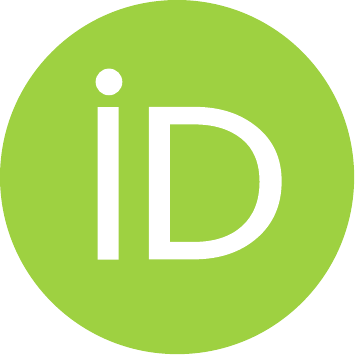}\hspace{1mm}Gustavo de Carvalho Bertoli}\\
    PhD Candidate\\
	Aeronautics Institute of Technology (ITA)\\
	São José dos Campos, SP, 12228-900, Brazil \\
	\texttt{gustavo.bertoli@ga.ita.br} \\
	\And
	\href{https://orcid.org/0000-0002-1568-9299}{\includegraphics[scale=0.06]{orcid.pdf}\hspace{1mm}Osamu Saotome} \\
	Department of Electronics Engineering\\
	Aeronautics Institute of Technology (ITA)\\
	São José dos Campos, SP, 12228-900, Brazil \\
	\texttt{osaotome@ita.br} \\
}


\maketitle


\begin{abstract}
This paper presents a Robot Operating System (ROS)/Gazebo application to calculate and simulate an optimal route for a drone in an urban environment by developing new ROS packages and executing them along with open-source tools. Firstly, the current regulations about UAS are presented to guide the building of the simulated environment, and multiple path planning algorithms are reviewed to guide the search method selection. After selecting the A* algorithm, both the 2D and 3D versions of them were implemented in this paper, with both Manhattan and Euclidean distances heuristics. The performance of these algorithms was evaluated considering the distance to be covered by the drone and the execution time of the route planning method, aiming to support algorithm’s choice based on the environment in which it will be applied. The algorithm execution time was 3.2 and 17.2 higher when using the Euclidean distance for the 2D and 3D A* algorithm, respectively. Along with the performance analysis of the algorithm, this paper is also the first step for building a complete UAS Traffic Management (UTM) system simulation using ROS and Gazebo.
\end{abstract}

\section{Introduction}
The increasing popularity of both civil and commercial Unmanned Aircraft Systems (UAS) applications has been noticeable in the past few years, as seen in the market growth for small commercial UAS. The FAA forecasts expect the sales of 2.7 million drones per year by 2020 \cite{utm07}. Recent technological advances allowed drones to be easier to operate, flexible, and relatively inexpensive, making them perfect for a wide range of applications, such as precision agriculture \cite{int02}, delivery of goods \cite{int03}, entertainment purposes, search and rescue missions, disaster relief, and infrastructure monitoring~\cite{int01}. 

The growing number of UAS required to meet the demand of their multiple applications represents a safety concern since the drones will share the airspace with general aviation. The presence of drones in restricted areas may cause accidents with catastrophic consequences, and flight delays and cancellations are not uncommon due to UAS flying near airports \cite{wikis}. New technologies must be developed to ensure all stakeholders' safety, privacy and integrity to enable commercial UAS operations. The UAS Traffic Management (UTM) is an important technology to guarantee the safety and organization of the airspace, and its concept of operations, developed by NASA, is presented by \cite{utm01}. 

This paper presents a platform that generates an optimal route to a UAS in a given simulated environment, starting with the extraction of the environment and decomposing it into cells, and then using the path planning algorithm to obtain an optimized route for the drone. Lastly, the drone navigation will be simulated considering the fixed constraints of the simulated environment. This paper is the starting point for building a complete UAS Traffic Management platform using the Robot Operating System (ROS), an open-source collection of software focused on developing robotics systems, and Gazebo, a software for systems simulation. The scientific community is increasingly applying the ROS and Gazebo tools, providing models and packages that are state-of-the-art in software development for robotics. A simulated UTM platform allows the performance analysis of the system and the anticipation of constraints in the early stages of the development, such as regulatory constraints, enabling the studies for the entry of commercial drones in the airspace \cite{bookros}. 

This paper is structured by the following sections: Section~\ref{sec2-utm} presents the literature review about the current efforts to develop UAS Traffic Management (UTM) systems. The development of regulations for UAS by ANAC, FAA, EASA and other governmental bodies all around the World is presented on Section~\ref{sec3-uas}. The main theoretical concepts about motion planning, considering environment representation, and path planning algorithms are presented by Section~\ref{sec-motion}. Section~\ref{sec:methods} describes the methodology for the development of this work, covering the building and analysis of the simulated environment, the chose and implementation of the route planning algorithm and the construction of the ROS package.
Section~\ref{sec:results} presents the results and discussion of the work, then, Section~\ref{sec:conclusion} presents the conclusion of the work and suggestions for future works.

\section{UAS Traffic Management}
\label{sec2-utm}

The entry of Unmanned Aircraft Systems (UAS) into the airspace is not a simple task. Since the number of UAS is rapidly increasing, the complexity of these aircraft's management also increases. The safety level for conventional aviation and civilians must be kept, even with hundreds of drones flying above the cities. Privacy is another concern with a high number of drones equipped with cameras. The environment and its constraints must be studied and defined to comply with these requirements, creating forbidden, restricted, and free areas for UAS. The operation process must be well defined, and counts with planning activities, execution management, and continuous improvement \cite{reg05}. This section presents an overview of the UTM systems under development in the USA and Europe.

\subsection{UTM in the US}

\begin{figure}[h]
    \centerline{\includegraphics[width=\textwidth]{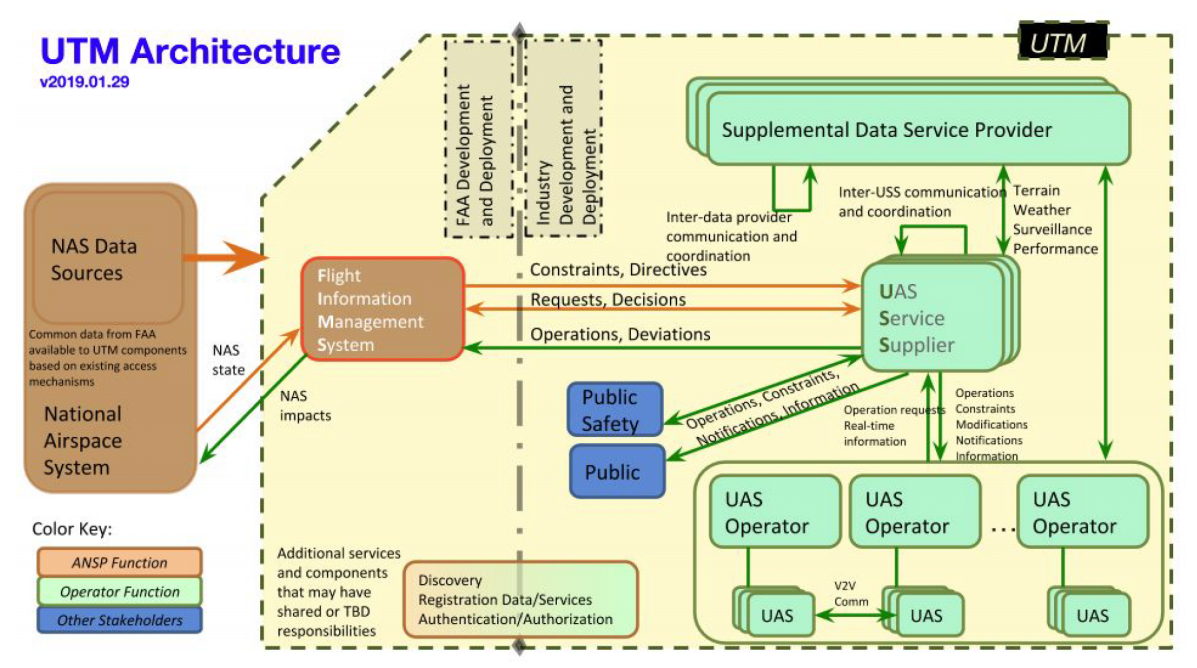}}
    \caption{Notional UTM architecture \cite{utmusa}} 
    \label{utmarch}
    \end{figure}

The UAS Traffic Management (UTM) system under development by Federal Aviation Administration (FAA) will support operations below 400 ft. A distributed information network will be a key component to achieve safe operations, allowing the data exchange and information sharing between FAA, operators, vehicles, and other stakeholders. The UTM will provide support for flight operations planning, communications, weather, and real-time constraints information, among others, to guarantee safe and secure operations. The proposed UTM architecture is represented in Figure~\ref{utmarch} \cite{utmusa}.

\subsection{U-Space in Europe}

The U-Space is a set of services presented by SESAR Joint Undertaking (SJU) and the European Aviation Safety Agency (EASA) to support the operation of commercial and non-commercial UAS in all types of airspace. The four sets of services to be implemented are \cite{uspace}:
\begin{itemize}
    \item U1: Fundamental services, such as electronic registration, electronic identification, and geofencing.
    
    \item U2: Initial services support, such as flight planning and approval, interface with air traffic control (ATC), among others.
    
    \item U3: Advanced services support, allowing operation for complex operations in high-density areas, such as automated detect and avoid and conflict detection.
    
    \item U4: Full services support, with a high level of automation, connectivity, and digitalization.
\end{itemize}

\section{UAS Regulations}
\label{sec3-uas}

The development of new regulations for UAS is getting more attention with the increasing expectations and according to forecasts for UAS deployment in coming years. These regulations and standards are still being discussed and changing since UAS technology is rapidly evolving. This section presents an overview of the regulations in Brazil and the rest of the world.

\subsection{National Civil Aviation Agency of Brazil}

The Brazilian Civil Aviation Special Regulation RBAC-E94 \cite{RBAC}, published by the National Civil Aviation Agency of Brazil (ANAC), addresses the general requirements for Unmanned Aircraft Systems. It considers this technology's current development stage to establish the operational conditions that guarantee civil aviation safety. This document forbids the autonomous operation of drones and establishes the pilots' requirements, such as minimum age, medical certification, and pilot license emitted by ANAC. Other legal requirements are discussed, such as UAS certificate, registration, and insurance for damage against third-parties. The Brazilian Department of Airspace Control (DECEA) issued the Aeronautics Command Instruction ICA 100-40 \cite{ICA100} to regulate procedures and responsibilities to guarantee safe access of UAS to the Brazilian airspace. The definition of general rules to access the airspace is one of the multiple contributions of this document, compiled in Table~\ref{tab_ica}.

\begin{table}[h]
\caption{General rules for UAS on Brazil \cite{ICA100}}
\label{tab_ica}
\centering
\begin{tabular}{|p{2.5cm}|p{3cm}|p{5cm}|} 
\hline
\multicolumn{1}{|c|}{\textbf{Altitude}} & \multicolumn{1}{c|}{\textbf{Minimum distance}} & \multicolumn{1}{c|}{\textbf{Constraint element}}                                \\ \hline
\multirow{2}{*}{0 to 400 ft}            & 30 m                                           & People not involved in operation, buildings, properties, structures and animals \\ \cline{2-3} 
                                        & No fly zone                                    & Above crowds and populated areas                                                \\ \hline
\multirow{5}{*}{0 to 131 ft}            & 5.6 km                                         & Registered aerodromes, operating on approach/takeoff zones                      \\ \cline{2-3} 
                                        & 1.9 km                                         & Registered aerodromes, operating outside of approach/takeoff zones              \\ \cline{2-3} 
                                        & 2 km                                           & Registered helipads with a height of up to 60 m                                 \\ \cline{2-3} 
                                        & 600 m                                          & Registered helipads with a height above 60 m                                    \\ \cline{2-3} 
                                        & 2 km                                           & Agricultural aviation areas                                                     \\ \hline
\multirow{3}{*}{131 to 400 ft}          & 9.3 km                                         & Registered aerodromes                                                           \\ \cline{2-3} 
                                        & 3 km                                           & Registered helipads                                                             \\ \cline{2-3} 
                                        & 2 km                                           & Agricultural aviation areas                                                     \\ \hline
\end{tabular}
\end{table}

\subsection{UAS Regulations from Around the World}

Several countries are committed to developing new regulations and standards for UAS, aiming to accompany this technology's accelerated growth. The current state regulations are accessible at the International Civil Aviation Organization's (ICAO) website \cite{icao_site}. One of the main elements that compose the regulation development process, along with pilot licenses, insurance, and registration, is the definition of restricted areas \cite{reficuas1}. Tables~\ref{tab_reg} and \ref{tab_reg2} presents a summary of safe distances between UAS and geofence elements from different state regulations \cite{reg05}. 

\begin{table}[h]
\caption{Minimum distance between UAS and geofence elements - Asia and Oceania \cite{reg05}}
\label{tab_reg2}
\centering
\begin{tabular}{|p{2.5cm}|p{3cm}|p{5cm}|}
\hline
\textbf{Country}                & \textbf{Minimum distance} & \textbf{Constraint element}                                                                \\ \hline
Australia                       & 30 m                                           & Person not directly associated with the operation                                                               \\ \hline

\multirow{5}{*}{China}          & 5 km                                           & National boundary lines, radio observations                                                                     \\ \cline{2-3} 
                                & 2 km                                           & Landing points for manned aircraft, borderlines, satellite earth stations                                       \\ \cline{2-3} 
                                & 1 km                                           & Military reservation, thunderhead, buildings, tall towers, power grids, wind power                              \\ \cline{2-3} 
                                & 500 m                                          & High-speed railway                                                                                              \\ \cline{2-3} 
                                & 200 m                                          & Warehouses with inflammable and explosive objects, petrol stations, electric power facilities, mountains        \\ \hline

\multirow{2}{*}{Japan}          & 30 m                                           & People and properties                                                                                           \\ \cline{2-3} 
                                & No-fly zone                                    & Over event sites with group of people                                                                           \\ \hline

\multirow{2}{*}{UAE}            & 5 km                                           & Airports, heliports, airfields and controlled airspace                                                          \\ \cline{2-3} 
                                & 150 m                                          & Crowds, public and private properties                                                                           \\ \hline
\end{tabular}
\end{table}

\begin{table}[h]
\caption{Minimum distance between UAS and geofence elements - Europe \cite{reg05}}
\label{tab_reg}
\centering
\begin{tabular}{|p{2.5cm}|p{3cm}|p{5cm}|}
\hline
\textbf{Country}                & \textbf{Minimum distance} & \textbf{Constraint element}                                                                \\ \hline
\multirow{3}{*}{Belgium}        & 2778 m                                         & Airports                                                                                                        \\ \cline{2-3} 
                                & 926 m                                          & Heliports                                                                                                       \\ \cline{2-3} 
                                & 50 m                                           & Buildings, people, animals                                                                                      \\ \hline
\multirow{2}{*}{Czech Republic} & 150 m                                          & Congested area                                                                                                  \\ \cline{2-3} 
                                & 100 m                                          & Person not directly associated with the operation                                                               \\ \hline
\multirow{3}{*}{Croatia}        & 3 km                                           & Airport and approach/departure zone                                                                             \\ \cline{2-3} 
                                & 150 m                                          & Group of people                                                                                                 \\ \cline{2-3} 
                                & 30 m                                           & People and structures                                                                                           \\ \hline
\multirow{2}{*}{Germany}        & 1.5 km                                         & Airports                                                                                                        \\ \cline{2-3} 
                                & No-fly zone                                    & Above people, accident and disaster areas, prisons, military installations, industrial areas and power stations \\ \hline
\multirow{3}{*}{Ireland}        & 8 km                                           & Airports                                                                                                        \\ \cline{2-3} 
                                & 2 km                                           & Aircraft in flight                                                                                              \\ \cline{2-3} 
                                & 150 m                                          & People, vessel, vehicle and structures                                                                          \\ \hline
\multirow{3}{*}{Italy}          & 5 km                                           & Airports                                                                                                        \\ \cline{2-3} 
                                & 150 m                                          & Congested areas                                                                                                 \\ \cline{2-3} 
                                & 50 m                                           & People and properties                                                                                           \\ \hline
Poland                          & 5 km                                           & Airports                                                                                                        \\ \hline
\multirow{2}{*}{Slovenia}       & 300 m                                          & Crowds                                                                                                          \\ \cline{2-3} 
                                & 50 m                                           & Powerlines, roads, railways, etc.                                                                               \\ \hline
Spain                           & 8 km                                           & Airports                                                                                                        \\ \hline
Sweden                          & 50 m                                           & People, animals and properties                                                                                  \\ \hline
\multirow{2}{*}{Switzerland}    & 5 km                                           & Airfields                                                                                                       \\ \cline{2-3} 
                                & 100 m                                          & Crowds                                                                                                          \\ \hline
\multirow{2}{*}{UK}             & 150 m                                          & Congested area                                                                                                  \\ \cline{2-3} 
                                & 50 m                                           & Person, object, vehicle                                                                                         \\ \hline
\end{tabular}
\end{table}                                                                     
\section{Motion Planning}
\label{sec-motion}
Motion planning, also referred to as path planning, aims to find a continuous free path from the start position to the desired goal pose. The identification of free and occupied spaces in the environment is mandatory to complete this objective. The environment may or may not be known a priori, and it may be static, not changing over time, or dynamic, with moving obstacles. The adoption of path planning algorithms is important for UTM development to create drone routes following restricted areas' regulatory constraints and safe distances. Path planning can solve many real-world problems, such as, robots navigation, VLSI (Very Large Scale Integration) layout, traveling salesman problem, and assembly sequencing.

Since multiple suitable paths between two points may be found by the path planning algorithms, it is important to define additional constraints to the problem in order to choose and optimal path \cite{pp12}. Some of these constraints are:

\begin{multicols}{2}
\begin{itemize}
    \item Shortest path length between start and goal positions;
    \item Shortest time to reach destination;
    \item Lowest fuel consumption;
    \item Higher safety level;
    \item Lowest collision risk.
\end{itemize}
\end{multicols}

\begin{figure}[h]
\centerline{\includegraphics[width=0.8\textwidth]{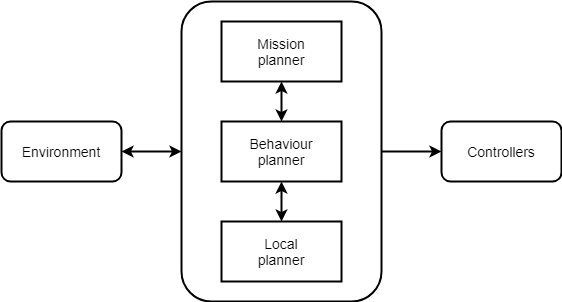}}
\caption{Motion planning hierarchical architecture}
\label{motion_planning}
\end{figure}

For complex applications, the motion planning problem might be decomposed in a hierarchical architecture \cite{ppn01} as shown in Figure~\ref{motion_planning}.

The mission planning task is responsible for the map-level navigation, focusing on the high-level constraints of the map, such as buildings, roads and terrain. Its output is the route from the starting point to the vehicle destination. The behavior planning focuses on the decision making based in the regulatory elements which rules the environment, such as traffic lights, signs, geofences and obstacles, among others. Its output is a sequence of high level control actions, constrained by sensor data, set of rules, etc. The local planning is responsible for following the path calculated by the mission planner, constrained by the behavior planner output and by the real-time sensor data. Its output is a feasible, collision-free and smooth trajectory that the vehicle shall follow \cite{ppn01}.

\subsection{Environment Representation}

\begin{figure}[ht]
\begin{minipage}{0.3\linewidth}
\includegraphics[width=\textwidth]{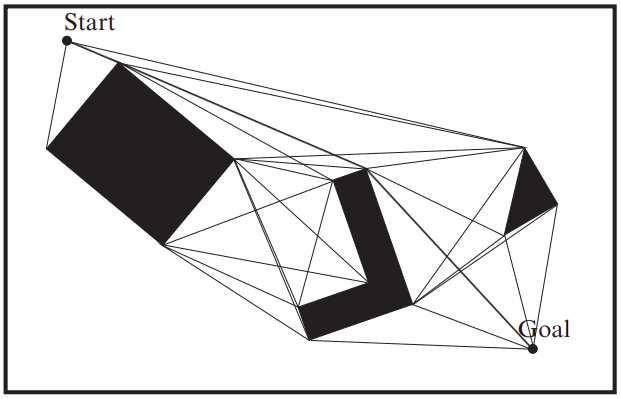}
\caption{Visibility graph \cite{pp12}}
\label{visi_graph}
\end{minipage}%
\hfill
\begin{minipage}{0.3\linewidth}
\includegraphics[width=\textwidth]{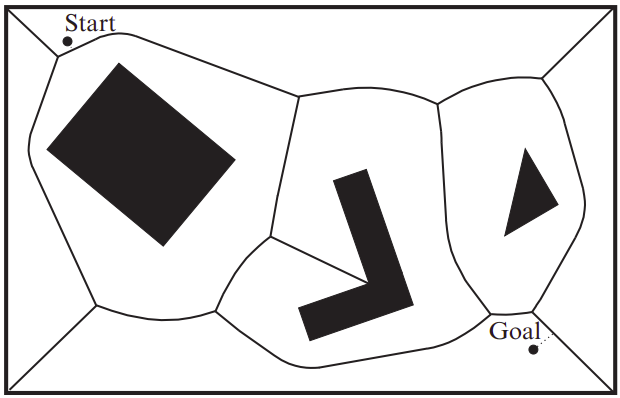}
\caption{Voronoi graph \cite{pp12}}
\label{voro_graph}
\end{minipage}%
\hfill
\begin{minipage}{0.3\linewidth}
\includegraphics[width=\textwidth]{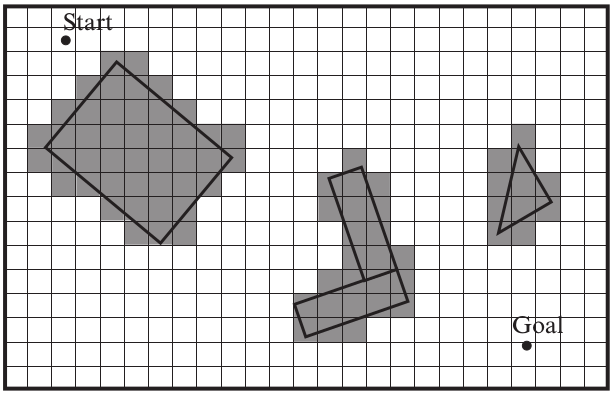}
\caption{Cell decomposition \cite{pp12}}
\label{cell_deco}
\end{minipage}
\end{figure}

An important part of the motion planning problem is the environment representation. Some of the ways of environment construction for path planning algorithms are:

\begin{itemize}
    \item Visibility graph: This representation, widely used in known environments for UAS applications, consists in the construction of undirected graphs between two points in the map, connected by the vertices of the obstacles, as shown in Figure~\ref{visi_graph}
    . Visibility graphs generate the shortest path \cite{env01}, but a  limitation of this representation is that its efficiency decreases as the number of obstacles increases.
    

    \item Voronoi graph: Consists in the generation of equidistant paths to the obstacles in the environment, as shown in Figure~\ref{voro_graph}
    . Although the generated path may not be optimal \cite{norvig}, this representation is indicated for safety-critical applications or for systems more susceptible to uncertainties in their localization measurements.
    


    \item Cell decomposition: This representation consists of dividing the environment into simple geometric shapes called cells \cite{pp12}. This decomposition can be \textit{accurate}, with each cell composed only by free or occupied spaces, or \textit{approximate}, when part of free space is defined as occupied by the decomposition algorithm, as shown in Figure~\ref{cell_deco}
    . In order to minimize the memory usage, the approximate cell decomposition may use \textit{variable cell size} \cite{env05}.
    
    
\end{itemize}

\subsection{Path Planning Algorithms}
\label{sec:algos}
Based on the representation of the environment, path planning algorithms are used for finding a feasible global path. There are many path planning algorithms, and choosing one for an application is not an easy task. In order to help the assessment of these methods, \cite{pp02} proposed the following taxonomy for 3D path planning methods:

\subsubsection{Sampling-based algorithms} 

    This category of algorithm does not require an explicit presentation of the environment. Some of the algorithms in this category can generate a graph or roadmap presentation and then find a feasible route between a set of starting and ending points. Since the presentation of a complex environment may have a high computational cost, this poses as an advantage of this category of algorithms. 
    
    The sampling-based algorithms select random points in the environment and then check if the path to reach it from the previous location is collision-free. The sampled random points and their connections in the free space might generate a path between the initial and desired positions.
    
    One sampling-based algorithm commonly used for UAS path planning applications is the Rapidly-exploring Random Tree (RRT). It builds a path between two points by sampling a new random node and connecting it to the closest point in the generated graph.
    
    
    For UAS applications, the RRT planner was implemented for navigation and exploration of unknown and cluttered environment \cite{rrt01}. A Predictive Rapid-exploring Random Tree algorithm was implemented for collision avoidance in dynamic environments \cite{rrt03}. For search and rescue missions, the RRT algorithm was able to avoid static obstacles and no-fly zones, generating a collision-free path \cite{rrt04}.
    
    
    The Probabilistic Roadmap (PRM) is a sampling-based algorithm that builds a presentation of the environment in order to enable the generation of feasible paths between different sets of initial and goal positions. A graph of multiple points and connections in the free space is generated in the \textit{learning phase}, and in the \textit{path searching phase} the closest possible points in the graph are selected to link two selected points on the map.

    
    A probabilistic roadmap based path planning algorithm was implemented for the generation of a collision-free path for a UAS in urban environments \cite{prm01}. The PRM algorithm was also implemented with other path planning algorithms, such as A* \cite{prm03} and D* \cite{prm02}.
    

\subsubsection{Node-based optimal algorithms}
    
    The node-based optimal algorithms take as an input the decomposed node-grid from the environment, with all the information about free and occupied spaces already processed. 
    
    The algorithms in this category repeatedly check if the current node (the first is the start point) is the goal node. If it is not, the current node is marked as already visited, and a neighbor point is selected to be the next current node. This process is repeated until the goal point is reached or after the evaluation of all nodes.
    
    The famous Dijkstra's algorithm falls into this category. It tracks the distance from the current node to the next evaluated node and adds it to the total path cost for a given path, providing an optimal route from starting to the goal position.    
    
    
    Dijkstra's algorithm was implemented for motion planning of a fixed-wing UAS based on terrain represented by digital elevations models \cite{dij02}. This algorithm was also used to solve the path planning problem considering the minimal gross propulsion energy \cite{dij01}. 
    
    
    The A* algorithm adds a heuristic function to the Dijkstra's algorithm to guide its search by estimating the distance from the next node to be evaluated to the goal position. This value enables the algorithm to guess the best next current node, providing an optimal path with better performance of the search.
    
    
    The A* algorithm was applied for UAS dynamic path planning in a mission of interception of moving targets \cite{pp14}. It was also implemented using a cost function that considers the distance to the target point and the required energy to accomplish the mission \cite{pp05}. Another application of the A* algorithm is path planning considering the wind in the environment \cite{astar02}.
    


\subsubsection{Mathematical model-based algorithms}
    
    The algorithms in this category describe the environment and the system as equations and inequalities, defining the kinematic and dynamic constraints of the path planning problem. The cost function finds an optimal solution considering these boundaries.
    
    The Mixed-Integer Linear Programming (MILP) methods combine discrete and linear variables to represent both system and environment, being able to model multiple kinds of problems, including path planning \cite{magatao}. 
    
    
    The MILP techniques were applied for the development of a UAS delivery service scheduling model \cite{milp02}. A path planning optimization using mixed-integer linear programming in Arctic environments was proposed in \cite{MESTRADO_ICE}, and a real-time trajectory planning in a dynamic environment was also implemented with MILP \cite{milp04}.
    

\subsubsection{Bio-inspired algorithms}

    The bio-inspired algorithms mimic the natural behavior of humans and other creatures to solve diverse computer science problems, including the path planning problem. \cite{pp08} presents an overview of the application of the algorithms in this category for robot path planning implementations.
    
    The Ant-Colony Optimization (ACO) algorithm mimics the behavior of ants in their search for food. While traveling through a path, the ant leaves a trail of pheromones that attracts other ants. The more pheromone in a trail, the higher the probability of the next animal chooses that path. After multiple iterations, the pheromones in the least used routes evaporate and are continuously deposited in the most used trail, being selected more often by the ants. By this, the ants can find the shortest path between the nest and the source of food.
    
    The algorithm mathematically models this behavior, calculating the probability of choosing each path based on the amount of pheromone deposited and evaporated in each trail. After multiple iterations, the result is the shortest path calculated between the start point and the goal node.
    
    
    The ant colony optmization was implemented to find an optimal route for UAS considering both the shortest path and detected hazards \cite{acor01}. An optimal global route for a UAS in a complex environment could also be found by an improved ant colony algorithm \cite{aco01}. An improved ant colony system algorithm for multi-obstacle path planning is presented in \cite{acor02}. The ant colony optimization also showed good results for UAS path planning in indoor environments \cite{aco03}.


    The inspiration for the Genetic Algorithm (GA) is Darwin's theory of natural evolution, mathematically modeling the process of natural selection, reproduction and genetic mutation.
    
    For the first iteration, the algorithm generates the first generation randomly and evaluates the fitness value of each solution, based on their chromosome values. The candidates with higher fitness values are more likely to be selected as "parents" for the next generation. The reproduction process is the random trade of the parents' chromosomes, and the genetic mutation occurs with the alteration of some values of the chromosomes in the new generation. This process iterates until the convergence criteria are met or when a predefined number of iterations is reached \cite{gen04}.
    
    
    Since the genetic algorithm has a high computational cost, the implementation of this algorithm for real-time UAS path planning using an FPGA (Field Programmable Gate Array) is presented in \cite{gen01}, and the use of a GPU (Graphics Processing Unit) for military UAS applications is shown in \cite{gen03}. A new algorithm called multi-frequency vibrational genetic algorithm was developed to reduce the computational time for UAS path planning \cite{gen02}.
    
    
    
    The Particle Swarm Optimization (PSO) is an algorithm inspired by bird flocking and fish schooling. When the animals are searching for food, they follow the one which is the closest to the food source. The algorithm implements this behavior by randomly distributing solutions (particles) across the solutions space and evaluating the fitness value of each particle. Then, two values are updated: The first one is the best fitness value that the particle had itself, and the second one is the best fitness value for all the particles. These values will determine the direction and velocity of the next movement of the particles.
    
    
    The particle swarm optimization algorithm was successfully implemented for a reconnaissance mission conducted by UAS swarms \cite{psor01}. An improved particle swarm optimization was studied to improve the rapidity and optimality of the path planning for UAS formation \cite{psor02}. Another implementation of the PSO in UAS path planning is for data gathering from a wide area Wireless Sensor Network \cite{pso04}. 
    
    
    
    
    
    

\section{Methodology} \label{sec:methods}

\begin{figure}[h]
\centerline{\includegraphics[width=1\textwidth]{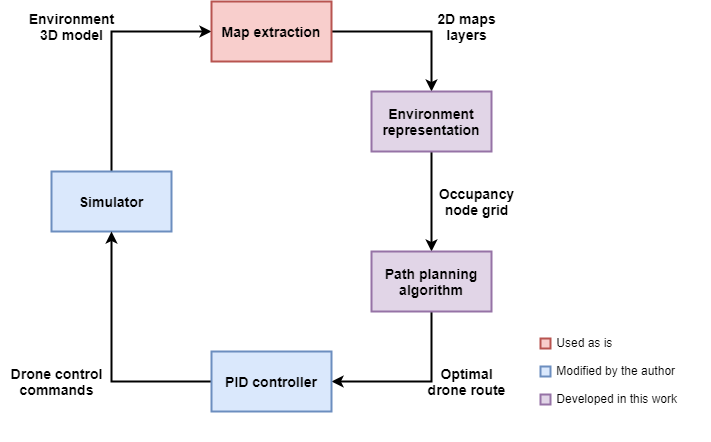}}
\caption{Software Block Diagram}
\label{sw_blocks}
\end{figure}

This section presents the methodology behind the development of the platform for path planning and simulation of a UAS in an urban environment. The tools developed in this paper are integrated with the software developed by other researchers to create the starting point of a complete UAS Traffic Management (UTM) platform. The software chosen for the implementation of this work was the Robot Operating System (ROS) and Gazebo, for programming and simulation, respectively. The ROS is an open-source collection of software frameworks focused on robotics development. Due to its modular approach, ROS support very well the code reuse, facilitating the access to the state-of-the-art in software development and its use. With its application becoming increasingly popular in the scientific community, it is possible to integrate different research results to develop new tools. Gazebo is an open-source robotics simulator with great integration with ROS, enabling the test of the algorithms implemented and the design of new robots in complex environments. Figure~\ref{sw_blocks} presents an overview of the tool development. Each block represents a step in the execution of the simulation and will be presented in depth in the following sections of this chapter.

\subsection{Simulator} \label{sec:met_sim}

\begin{figure}[h]
\centerline{\includegraphics[width=0.85\textwidth]{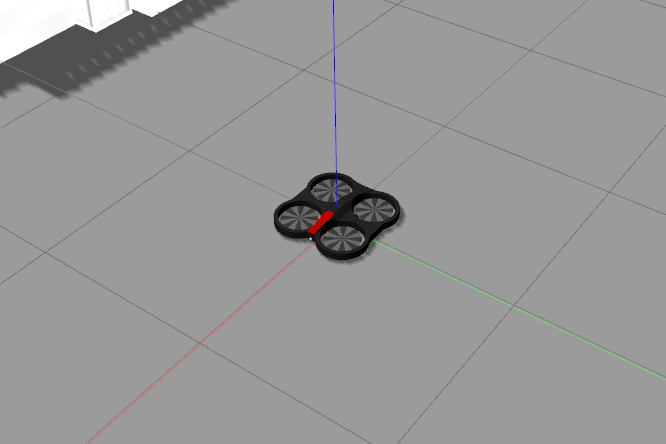}}
\caption{Parrot AR.Drone 2.0 model at \textit{tum\_simulator}}
\label{ardrone}
\end{figure}

The gazebo simulation of the 3D environment and the drone model is implemented by the ROS package \textit{tum\_simulator}\footnote{https://github.com/angelsantamaria/tum\_simulator}, developed by \cite{tum_sim} of the Computer Vision Group at the Technical University of Munich. The quadcopter model used in this simulator, the Parrot AR.Drone 2.0, is shown in Figure~\ref{ardrone}.

\begin{figure}[h]
\centerline{\includegraphics[width=0.75\textwidth]{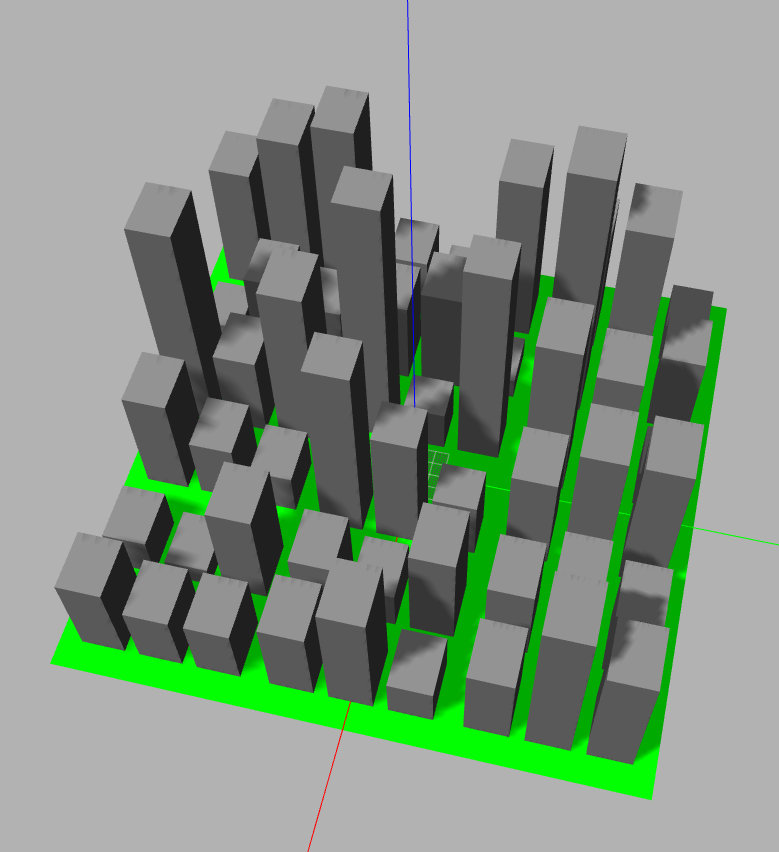}}
\caption{Simulation of an urban environment on Gazebo}
\label{cityenv}
\end{figure}

For the development of this paper, the environment created with the Gazebo simulation tool represents a dense urban environment, as shown in Figure~\ref{cityenv}. This kind of environment allows the observation of the UAS behavior in accordance with the separation rules defined by the regulatory agencies due to safety, security, integrity, privacy, and noise concerns.

The central court in this simulated environment have a reserved area for take-off and landing activities. The buildings in this region have dimensions of 15 x 15 meters, while the buildings in the surrounding courts have dimensions of 20 x 15 meters. Their heights were randomly defined, between 10 and 120 meters.


\subsection{Map Extraction} \label{sec:met_map}

\begin{figure}[h]
\centerline{\includegraphics[width=0.65\textwidth]{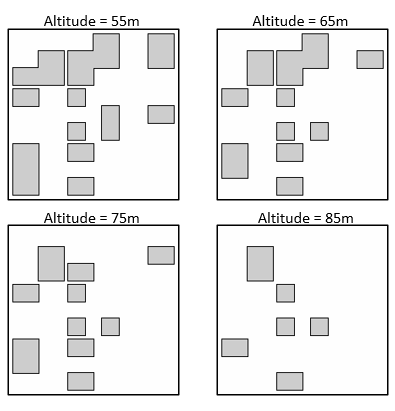}}
\caption{Two-dimensional maps generated for multiple heights.}
\label{2dmaps}
\end{figure}

The environment representation selected for this paper was the cell decomposition to be used with the A* algorithm. In order to build the occupancy grid for the use of path planning algorithms, multiple two-dimensional maps layers for different heights were extracted from the simulated environment. These heights represent the flight levels where the UASs are expected to operate. The software tool used to build these 2D maps was the \textit{gazebo\_ros\_2Dmap\_plugin}\footnote{https://github.com/marinaKollmitz/gazebo\_ros\_2Dmap\_plugin} \cite{gzbplugin}. This plugin performs a wavefront exploration to automatically generate a two-dimensional occupancy map from the simulated worlds in Gazebo. Figure~\ref{2dmaps} presents the maps generated for each height from the environment built for this work.

The outputs of the plugin are portable gray map files, and their resolution is defined by the user. For this paper, each pixel represents one square meter in the environment.


\subsection{Environment Representation} \label{sec:met_env}

\begin{figure}[h]
\centerline{\includegraphics[width=0.75\textwidth]{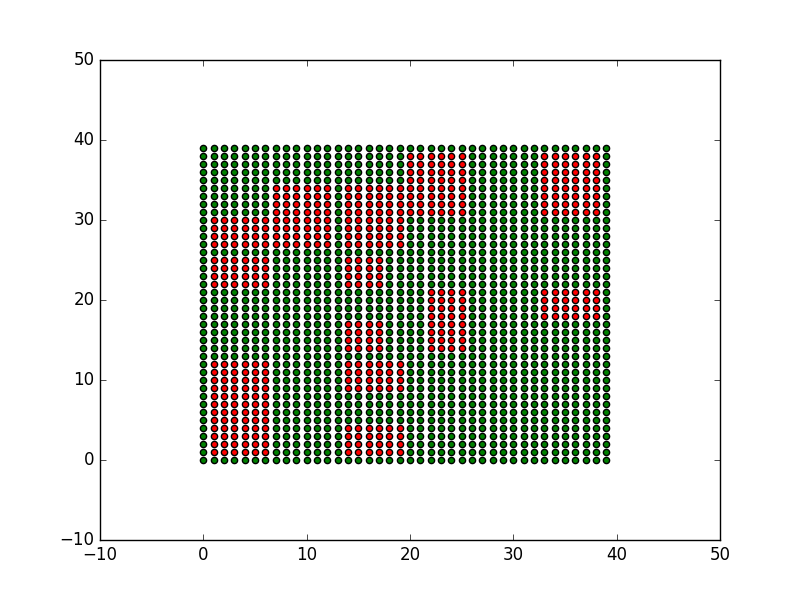}}
\caption{Two-dimensional node grid representation.}
\label{node_grid_2d}
\end{figure}

To represent the simulated environment in a way that allows the application of the path planning algorithm, a tool was developed to generate a three-dimensional occupancy grid based on the two-dimensional maps generated by the Gazebo plugin. The first step performed by this tool is the conversion of each map image into a two-dimension numeric matrix, where each pixel is evaluated to get its grayscale value, which is placed in the matrix with the respective x and y value. The free points are represented as white points, so their grayscale values are equal or higher than 254. The grey or black points represent the occupied spaces, so their grayscale values are less than 254. The two-dimensions matrices are then stacked in order, generating a three-dimension array with the values. Figure~\ref{node_grid_2d} presents the 2D node grid generated from the extracted 2D map.


\begin{figure}[h]
\centerline{\includegraphics[width=0.85\textwidth]{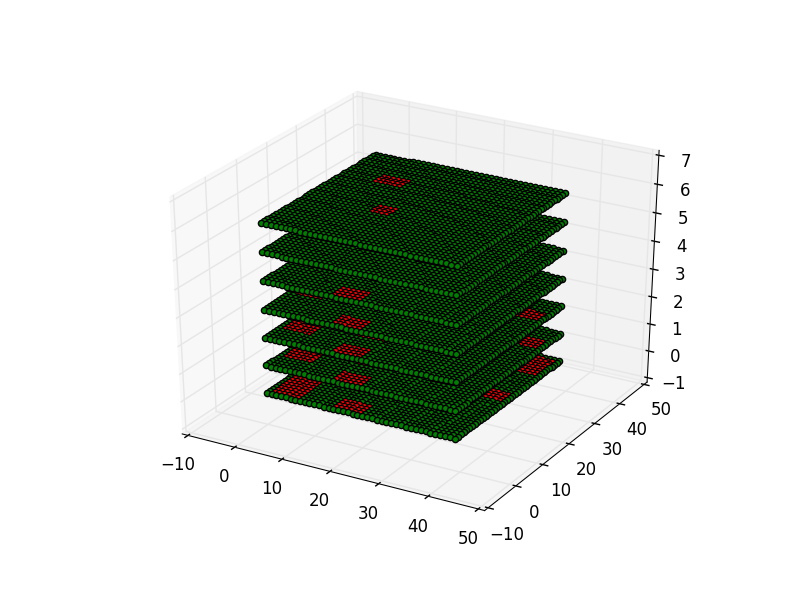}}
\caption{Three-dimensional node grid visual representation.}
\label{node_grid_3d}
\end{figure}

The following step is the definition of restricted areas due to regulatory constraints. Due to safety, security, privacy, noise, and other concerns, a safe distance between buildings and drones must be respected, and the UAS regulations across the world are responsible to define them. Since these constraints are not well defined and they vary from country to country, they were left as an input for the user, allowing the test for different configurations. The tool checks each value of the matrix to determine if that position is a free or occupied point. If that is an occupied point, all the points within the safety margin will be classified as a restricted area, represented in the final node grid as one. The free spaces are represented as zero. The last and optional step of the node grid generation tool is the reduction of the maps' scale by grouping the matrix elements in 5 x 5 sets. If any of the elements in the set represented an occupied space, the whole set was classified as an occupied node. For large environments, this step might be required in order to reduce the number of nodes to be visited by the path planning algorithm, reducing its execution time. On the other hand, this might lead to a loss of detail, which may hide some narrow but feasible paths for the drone. A visual representation of the node grid is shown in Figure~\ref{node_grid_3d}, with the green points representing the free spaces and the red points representing the occupied nodes.

\subsection{Path Planning Algorithm} \label{sec:met_pp}


The next step for the development of this paper was selecting the path planning algorithm to be implemented. As seen in section~\ref{sec:algos}, there are plenty of methods to solve the route planning problem, and multiple criteria can determine the best one to be chosen. The main requirements for the planner are:

\begin{itemize}
    \item Static threats handling, since the tool only handles the mission planning phase
    \item Generation of an optimal path
    \item Fast searching ability, to reduce the computational cost
\end{itemize}

An algorithm that suits these requirements is the A* algorithm, outperforming algorithms such as the RRT \cite{RW02} and the ant-colony optimization \cite{astar03}. Another advantage is that the A* is a deterministic algorithm, and the use of non-deterministic software in safety-critical systems is still a big challenge for certification \cite{DET01}. 

The A* algorithm starts with the creation of two lists: A list for the nodes that may constitute the optimal path and might be evaluated, that we call \textit{OpenNodes} list, and a list for the nodes that have already been evaluated, the \textit{ClosedNodes} list. The cost function of the nodes on \textit{OpenNodes} is then calculated, based on the traveled distance from the starting point, and the estimate of distance to the end point calculated by a heuristic function. The total traveled distance is the sum of the Euclidean distances between each node of the path with its successor. The heuristic function, on the other hand, is the Euclidean or Manhattan distance between the current node and the goal point.

\begin{figure}[h]
\centerline{\includegraphics[width=0.75\textwidth]{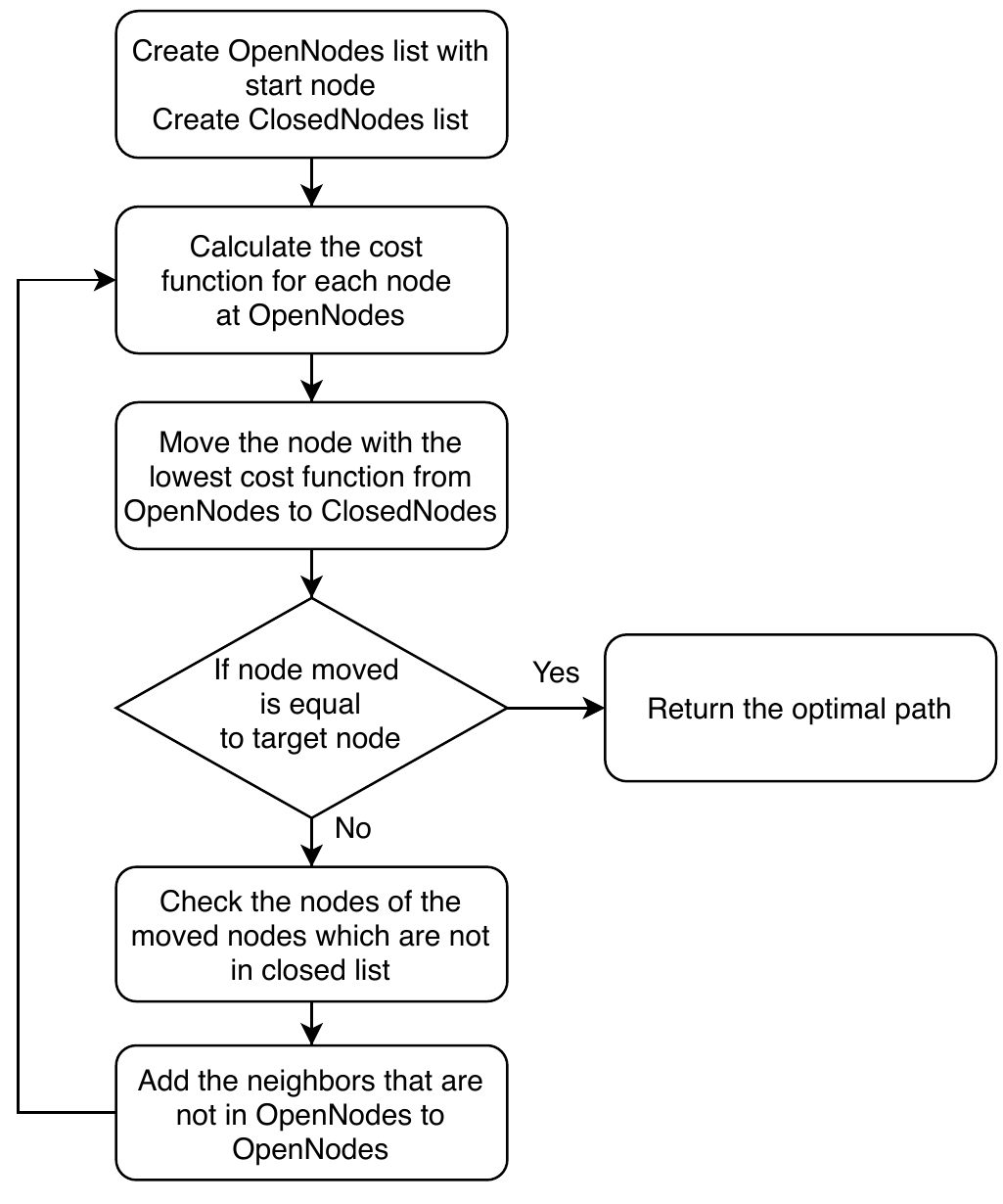}}
\caption{A* algorithm flowchart.}
\label{astarfc}
\end{figure}

The node with the lowest cost function is moved from the \textit{OpenNodes} list to the \textit{ClosedNodes} list, and if it is the goal node, the algorithm is terminated and returns the optimal path. Otherwise, the neighbors nodes of the current node selected are added to the \textit{OpenNodes} list and the algorithm is repeated until the target point is reached or until all the nodes are evaluated and no feasible path is found. Figure~\ref{astarfc} presents the algorithm flowchart.

\subsection{PID Controller} \label{sec:met_pid}

Once the optimal path is generated by the path planning algorithm, the simulation of the drone following this path is executed. The nodes coordinates are converted to the Gazebo map coordinates and then the Proportional Integral Derivative (PID) Controller\footnote{https://github.com/carlospotter/pid\_control\_ardrone} sends the commands to the drone in the simulated environment. This tool was originally developed by \cite{pid_git} and a few modifications have been made for the application in this work. The first modification was adding the landing command to the drone when reaching the final destination. Originally, the UAS would go straight back to the starting point after reaching its goal, without avoiding the obstacles as desired for our UTM implementation. Some values have also changed in order to comply with the capabilities of the simulator. The position and angle error set-points increased compared to the original configuration of the tool due to the accuracy of the pose measurement of the simulator. 


\section{Results and Discussion}
\label{sec:results}

\subsection{UAS Simulation}

\begin{figure}[h]
\centerline{\includegraphics[width=0.8\textwidth]{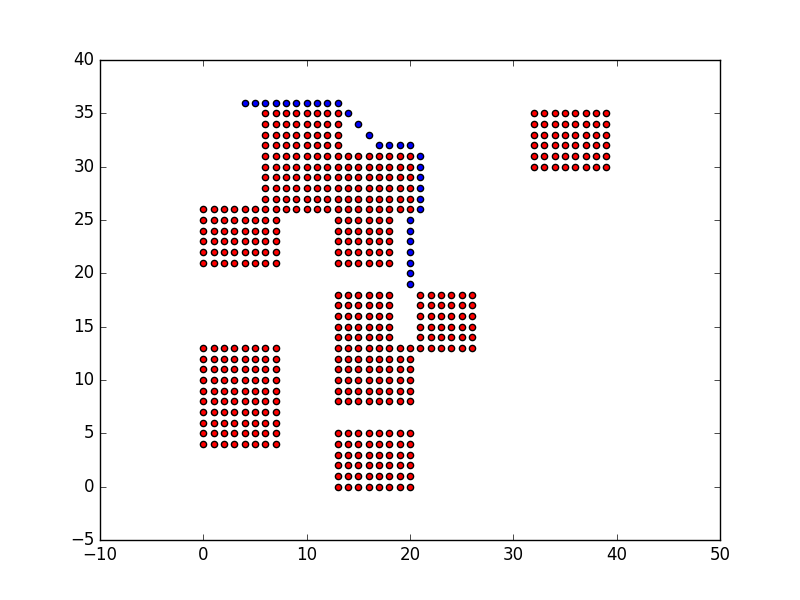}}
\caption{2D route calculated using A* algorithm}
\label{2drt}
\end{figure}

The package developed in this work enabled the map representation, path planning and simulation of the UAS in the Gazebo environment, considering fixed constraints. Figure~\ref{2drt} presents the calculated route for the 2D algorithm from the start position to one of the landing sites specified in the simulated environment.

The UAS then travel the previously calculated route in the simulated environment in a fixed altitude, as presented in a video demonstration\footnote{https://youtu.be/aGxPD6Bk608}. The execution of the 3D A* algorithm allows the drone to change its altitude in order to achieve even shorter paths, and is also demonstrated in the video.

This platform is the starting point of the complete UAS Traffic Management simulation platform. It is reproducible and can be used and modified to add new functionalities and capabilities. 

\subsection{2D and 3D A* Algorithm Comparison}

This section presents the performance comparison between the 2D and 3D A* algorithm implemented in the development of this work, using the Euclidean distance heuristic function. For the trade-off analysis were considered both execution time of the algorithm and total distance calculated for the optimal route, as presented in \cite{icuasita}.

\begin{figure}[h]
\centerline{\includegraphics[width=0.75\textwidth]{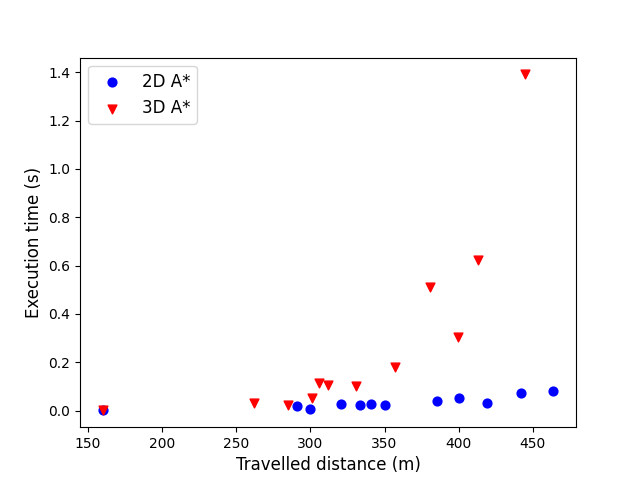}}
\caption{Comparison between 2D and 3D A* algorithm with Euclidean distance heuristics}
\label{img_icuas}
\end{figure}

Table~\ref{tab_icuas} presents the optimal calculated route length and the execution time for both 2D and 3D A* algorithm implementations. The results are also presented in figure~\ref{img_icuas}. 

\begin{table}[h]
\caption{2D and 3D A* algorithm comparison}
\label{tab_icuas}
\centering
\begin{tabular}{|c|c|c|c|c|c|}
\hline
\textbf{Start} & \textbf{Goal}    & \textbf{3D cost (m)} & \textbf{2D cost (m)} & \textbf{3D time (ms)} & \textbf{2D time (ms)} \\
\hline
(0,0,0) & (0,0,10)  & 160     & 160     & 0.44    & 1.39   \\
(0,0,0) & (0,0,30)  & 284.72  & 300     & 22.14   & 7.26   \\
(0,0,0) & (0,10,30) & 301.42  & 320.71  & 50.33   & 25.38  \\
(0,0,0) & (0,20,10) & 261.77  & 290.71  & 31.16   & 17.3   \\
(0,0,0) & (0,20,20) & 305.91  & 333.64  & 113.83  & 22.6   \\
(0,0,0) & (0,20,30) & 330.91  & 350.21  & 99.46   & 22.35  \\
(0,0,0) & (0,30,10) & 311.77  & 340.71  & 104.54  & 28.28  \\
(0,0,0) & (0,30,30) & 380.91  & 400.21  & 509.52  & 50.06  \\
(0,0,0) & (0,30,39) & 399.55  & 418.85  & 303.67  & 33.14  \\
(0,0,0) & (0,39,10) & 356.77  & 385.71  & 181.76  & 37.93  \\
(0,0,0) & (0,39,30) & 412.83  & 441.78  & 621.11  & 70.55  \\
(0,0,0) & (0,39,39) & 444.55  & 463.85  & 1391.44 & 79.08  \\
\hline
\end{tabular}
\end{table}

For this scenario, the traveled distance calculated by the 3D algorithm is about 6\% better than the one calculated by the 2D A*, with an execution time nine times higher, on average.

\subsection{Heuristic Functions Performance Comparison}

The comparison between the 2D and 3D A* algorithm was then extended to quantify the influence of the heuristic function selected for the path planning algorithm in the execution time and length of the the route calculated.

The selection of the heuristic function may control the behaviour of the A* algorithm \cite{theostf}. When the heuristic function is always lower or equal to the actual cost of moving to the goal position, the A* algorithm will provide the shortest path, but the software execution is slower. This is the case for the Euclidean distance. On the other hand, when the heuristic function sometimes returns a value greater than the actual cost of moving to the end point, the route calculated by the A* algorithm is not guaranteed to be the shortest, but the algorithm will run faster. The Manhattan distance is an example of heuristic with this behaviour. 


Tables~\ref{distancecomp} and \ref{runtimecomp} presents the comparison between the two implemented heuristics for both two and three-dimensional scenarios. For the 2D algorithm, the length of the routes calculated is the same for both heuristics, while the 3D routes calculated with the Manhattan heuristic were only 0.62\% longer, in average, than the routes calculated using the Euclidean distance, affecting only certain points. The execution time was 3.2 and 17.2 higher when using the euclidean distance for the 2D and 3D A* algorithm, respectively.

\begin{table}[h!]
\caption{Travelled distance comparison}
\label{distancecomp}
\centering
\begin{tabular}{|c|c|c|c|c|c|}
\hline
\multirow{2}{*}{Start} & \multirow{2}{*}{Goal} & \multicolumn{2}{c|}{2D Route Length (m)}                   & \multicolumn{2}{c|}{3D Route Length (m)}                   \\ \cline{3-6} 
                       &                       & Euclidean             & \multicolumn{1}{c|}{Manhattan} & Euclidean             & \multicolumn{1}{c|}{Manhattan} \\ \hline
(0,0,0)	& (0,0,10) & 160.0 & 160.0 & 160.0 & 160.0 \\ \hline 
(0,0,0)	& (0,0,30) & 300.0 & 300.0 & 284.72 & 284.72 \\ \hline 
(0,0,0)	& (0,10,30) & 320.71 & 320.71 & 301.42 & 302.42 \\ \hline 
(0,0,0)	& (0,20,10) & 290.71 & 290.71 & 261.77 & 261.77 \\ \hline 
(0,0,0)	& (0,20,20) & 333.64 & 333.64 & 305.91 & 316.35 \\ \hline 
(0,0,0)	& (0,20,30) & 350.21 & 350.21 & 330.91 & 330.91 \\ \hline 
(0,0,0)	& (0,30,10) & 340.71 & 340.71 & 311.77 & 311.77 \\ \hline 
(0,0,0)	& (0,30,30) & 400.21 & 400.21 & 380.91 & 380.91 \\ \hline 
(0,0,0)	& (0,30,39) & 418.85 & 418.85 & 399.55 & 399.55 \\ \hline 
(0,0,0)	& (0,39,10) & 385.71 & 385.71 & 356.77 & 356.77 \\ \hline 
(0,0,0)	& (0,39,30) & 441.78 & 441.78 & 412.83 & 425.91 \\ \hline 
(0,0,0)	& (0,39,39) & 463.85 & 463.85 & 444.55 & 444.55 \\ \hline 
\end{tabular}
\end{table}

\begin{table}[h!]
\caption{Execution time comparison}
\label{runtimecomp}
\centering
\begin{tabular}{|c|c|c|c|c|c|}
\hline
\multirow{2}{*}{Start} & \multirow{2}{*}{Goal} & \multicolumn{2}{c|}{2D A* execution time (ms)}                   & \multicolumn{2}{c|}{3D A* execution time (ms)}                   \\ \cline{3-6} 
                       &                       & Euclidean             & \multicolumn{1}{c|}{Manhattan} & Euclidean             & \multicolumn{1}{c|}{Manhattan} \\ \hline
(0,0,0)	& (0,0,10) & 1.39 & 1.25 & 0.44 & 0.45 \\ \hline 
(0,0,0)	& (0,0,30) & 7.26 & 6.63 & 22.14 & 9.57 \\ \hline 
(0,0,0)	& (0,10,30) & 25.38 & 9.98 & 50.33 & 17.12 \\ \hline 
(0,0,0)	& (0,20,10) & 17.3 & 10.61 & 31.16 & 18.67 \\ \hline 
(0,0,0)	& (0,20,20) & 22.6 & 10.58 & 113.83 & 18.5 \\ \hline 
(0,0,0)	& (0,20,30) & 22.35 & 10.24 & 99.46 & 17.77 \\ \hline 
(0,0,0)	& (0,30,10) & 28.28 & 11.97 & 104.54 & 19.24 \\ \hline 
(0,0,0)	& (0,30,30) & 50.06 & 11.51 & 509.52 & 18.89 \\ \hline 
(0,0,0)	& (0,30,39) & 33.14 & 12.53 & 303.67 & 18.67 \\ \hline 
(0,0,0)	& (0,39,10) & 37.93 & 13.24 & 181.76 & 20.35 \\ \hline 
(0,0,0)	& (0,39,30) & 70.55 & 12.62 & 621.11 & 19.88 \\ \hline 
(0,0,0)	& (0,39,39) & 79.08 & 13.2 & 1391.44 & 19.74 \\ \hline 
\end{tabular}
\end{table}

\begin{figure}[h!]
\centerline{\includegraphics[width=0.75\textwidth]{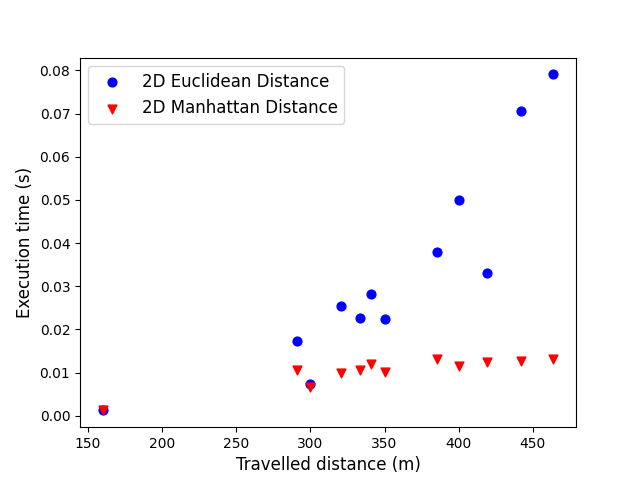}}
\caption{Comparison between different heuristic functions in 2D A* algorithm}
\label{2dres}
\end{figure}

\begin{figure}[h!]
\centerline{\includegraphics[width=0.75\textwidth]{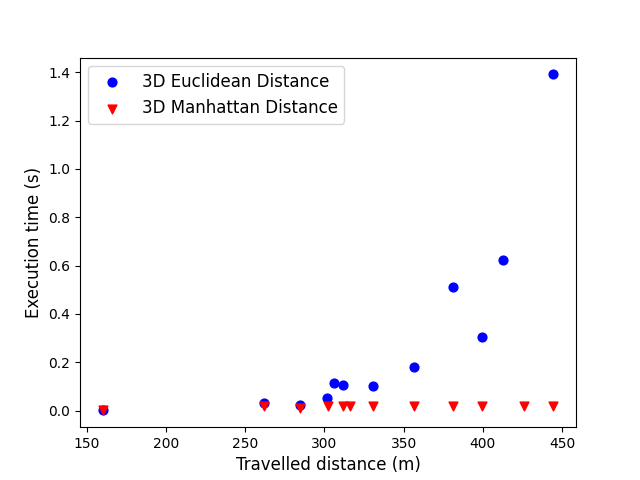}}
\caption{Comparison between different heuristic functions in 3D A* algorithm}
\label{3dres}
\end{figure}

Figures~\ref{2dres} and \ref{3dres} present how the execution time tend to increase with the calculated route length when using the Euclidean distance as heuristic function and remains nearly constant when using the Manhattan distance. This information is important in order to choose the more adequate path plannning method considering the size of the map, which may be larger than the one used in this work considering the application in an UTM system.

\subsection{Effect of the Map Representation in A* Algorithm Performance}
The map representation used in the work groups the nodes in 5x5 sets in order to reduce the complexity of the path planning algorithm. A drawback of increasing the granularity of the grids is the loss of detail in the map, so narrow feasible paths may not be identified by the path planning algorithm, leading to longer routes for the UAS.

Table~\ref{repcomp} presents a comparison between the routes and the execution times of the A* algorithm with the reduced map and with the original map as input. The execution time for the original map is about 85 times higher than when using the reduced map, in average, while the route length is about 8\% smaller.

\begin{table}[h!]
\caption{Algorithm comparison between reduced and original map representations}
\label{repcomp}
\centering
\begin{tabular}{|c|c|c|c|c|c|}
\hline
\multirow{2}{*}{Start} & \multirow{2}{*}{Goal} & \multicolumn{2}{c|}{Route Length (m)}                   & \multicolumn{2}{c|}{Execution Time (ms)}                   \\ \cline{3-6} 
                       &                       & Reduced map             & \multicolumn{1}{c|}{Original map} & Reduced map             & \multicolumn{1}{c|}{Original map} \\ \hline
(0,0,0)	& (0,0,10) & 160.0 & 162.83 & 1.39 & 16.64 \\ \hline 
(0,0,0)	& (0,0,30) & 300.0 & 262.83 & 7.26 & 118.99 \\ \hline 
(0,0,0)	& (0,10,30) & 320.71 & 300.53 & 25.38 & 1411.0 \\ \hline 
(0,0,0)	& (0,20,10) & 290.71 & 256.38 & 17.3 & 698.02 \\ \hline 
(0,0,0)	& (0,20,20) & 333.64 & 295.25 & 22.6 & 1249.74 \\ \hline 
(0,0,0)	& (0,20,30) & 350.21 & 349.52 & 22.35 & 1884.82 \\ \hline 
(0,0,0)	& (0,30,10) & 340.71 & 283.54 & 28.28 & 1298.9 \\ \hline 
(0,0,0)	& (0,30,30) & 400.21 & 372.99 & 50.06 & 4141.03 \\ \hline 
(0,0,0)	& (0,30,39) & 418.85 & 417.35 & 33.14 & 3896.39 \\ \hline 
(0,0,0)	& (0,39,10) & 385.71 & 328.54 & 37.93 & 2042.75 \\ \hline 
(0,0,0)	& (0,39,30) & 441.78 & 395.74 & 70.55 & 7763.53 \\ \hline 
(0,0,0)	& (0,39,39) & 463.85 & 436.63 & 79.08 & 8893.14 \\ \hline

\end{tabular}
\end{table}

\begin{figure}[h!]
\centerline{\includegraphics[width=0.75\textwidth]{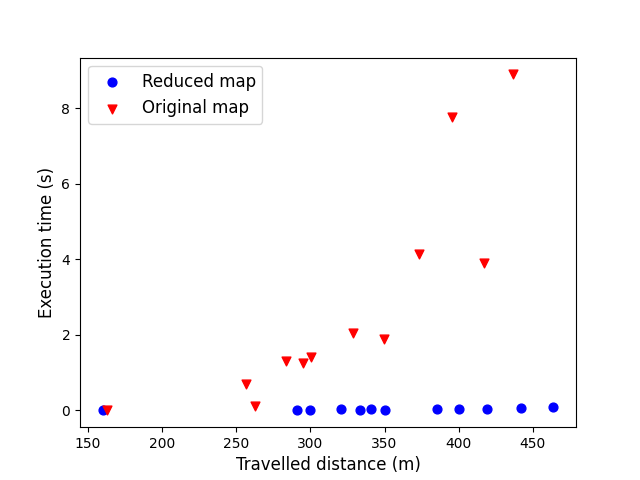}}
\caption{Comparison between A* algorithm using reduced and original map}
\label{repres}
\end{figure}

Figure~\ref{repres} presents how the execution time increases with the route length, pointing the necessity of reduce the size of the nodegrid for the path planning algorithm, specially when dealing with city maps for UTM systems.

\subsection{Statistical Tests for Algorithms Comparison}

To ensure greater accuracy in the performance comparison between the algorithms, measured by the execution time, and to ensure that the results were not random, a statistical test was performed.

For the statistical tests, the null hypothesis ($H0$) states that there are no differences between the algorithms, and it can be rejected with a \textit{p}-value $< 0.05$. The Wilcoxon signed ranks test was chosen for our comparison \cite{demsar}, and the results are represented in table~\ref{wcxres}.

\begin{table}[h!]
\caption{Wilcoxon test results}
\label{wcxres}
\centering
\begin{tabular}{|l|l|}
\hline
\multicolumn{1}{|c|}{\textbf{Comparison}}     & \multicolumn{1}{c|}{\textbf{p-value}} \\ \hline
2D A* with Euclidean and Manhattan heuristics & 0.000488                              \\ \hline
3D A* with Euclidean and Manhattan heuristics & 0.000977                              \\ \hline
2D vs 3D A* with Euclidean heuristic          & 0.000977                              \\ \hline
2D vs 3D A* with Manhattan heuristic          & 0.000977                              \\ \hline
2D A* with and without map reduction          & 0.000488                              \\ \hline
\end{tabular}
\end{table}

\section{Conclusion}
\label{sec:conclusion}
This paper presented a UAS Traffic Management system simulator. The tools developed for this paper are integrated to other previously published ROS packages to calculate an optimal route for the UAS and then simulate its trajectory on the simulated environment. A map representation is generated based on the occupancy map extracted from the simulated environment. The minimum distance allowed between the UAS and the obstacles is then defined based on the user input, to satisfy the desired regulation.

Two versions of the A* algorithm were implemented in this paper: one generating three-dimensional routes for the drone, and other generating two-dimensional paths for different altitudes and picking the shortest among them. 

The influence of the selection of the heuristic function in the A* algorithm was also analyzed, with the use of Euclidean and Manhattan distances in the path planning algorithm. The two versions of the A* algorithm were then compared using both heuristic functions. For the same start and goal points, the routes generated and the execution time were compared, with clear advantage of the Manhattan distance, specially when longer routes were required. 

The effect of the map representation in the A* algorithm performance was also studied in this work. The path planning algorithm was executed using a map representation with the nodes grouped in 5x5 sets and using the original map, with a 40x40 node grid versus a 200x200. The execution time for the second case was 84.5 times higher, in average.

Both analysis are important during the selection and implementation of the path planning algorithm, specially when dealing with large city maps.

This paper intends to be the starting point of the development of a complete UAS Traffic Management simulation platform. The future works based on this paper are the implementation of different path planning algorithms to evaluate their performance; generation of fixed air routes to comply with different regulations and optimize the path planning process; implementation of a schedule algorithm to handle multiple drones path planning; consider dynamic constraints in the path planning algorithm, allowing the simulation of multiple drones.



\section{Code availability}
The simulation data that support the findings and reproducibility of this study are available in GitHub / Zenodo with the identifier http://doi.org/10.5281/zenodo.4609280 




\bibliographystyle{abbrvnat}
\bibliography{bibfile}
\end{document}